\documentclass[runningheads]{llncs}

\usepackage{eccv}

\usepackage{eccvabbrv}

\newcommand{\methodname}{GARDEN\xspace}

\renewcommand{\paragraph}[1]{\vspace{.5em}\noindent\textbf{#1.}}

\usepackage{xspace}
\usepackage{soul}
\setuldepth{foobar}

\usepackage{graphicx}
\usepackage{booktabs}
\usepackage[table]{xcolor}
\usepackage{multirow}
\usepackage{multicol}
\usepackage{wrapfig}
\usepackage{placeins}

\usepackage[accsupp]{axessibility}  % Improves PDF readability for those with disabilities.

\usepackage{hyperref}

\usepackage{orcidlink}
\usepackage{marvosym}
\newcommand{\corrauth}{\textsuperscript{\textrm{\Letter}}}

\begin{document}

\title{\textsc{GARDEN}: Gravity-Aligned Reconstruction of Disentangled ENvironments from RGB images}

\titlerunning{\textsc{GARDEN}}

\author{Jiahao Sun\inst{1,2}\thanks{Work done while J. Sun, D. Wei, and H. Zhou were research interns at Ant Group.} \and
Dingkun Wei\inst{1,2}\textsuperscript{\ensuremath{\star}} \and
Zehong Shen\inst{2}\textsuperscript{\ensuremath{\dagger}} \and
Hongyu Zhou\inst{1,2}\textsuperscript{\ensuremath{\star}} \and \\
Yujun Shen\inst{2}\corrauth \and 
Liang Li\inst{1}\corrauth
}

\authorrunning{J. Sun et al.}

\institute{Zhejiang University,  China \and
Ant Group, China
}

\maketitle

% 1. 通讯作者小信封脚注
\begingroup 
\renewcommand{\thefootnote}{\textrm{\Letter}} 
\footnotetext{Corresponding author.} 
\endgroup

% 2. Project lead 脚注
\begingroup 
\renewcommand{\thefootnote}{\ensuremath{\dagger}} 
\footnotetext{Project lead.} 
\endgroup

%%%%%%%%% ABSTRACT
% \begin{abstract}
%   The abstract should concisely summarize the contents of the paper. 
%   While there is no fixed length restriction for the abstract, it is recommended to limit your abstract to approximately 150 words.
%   Please include keywords as in the example below. 
%   This is required for papers in LNCS proceedings.
%   \keywords{First keyword \and Second keyword \and Third keyword}
% \end{abstract}

\begin{abstract}
Converting multi-view RGB observations into simulation-ready 3D environments remains challenging because current reconstruction pipelines produce monolithic scene representations without explicit physical structure. 
They are typically defined up to an arbitrary global rotation and entangle rigid foreground objects with background geometry, which hinders stable physical interaction. 
Existing solutions often recover interactivity by replacing reconstructed objects with retrieved CAD assets, but this introduces a slow retrieval-and-replacement stage and weakens scene-specific geometric fidelity. 
We propose \methodname, an RGB-only framework that reformulates reconstruction as \textit{physically-grounded scene factorization} and outputs a structured hybrid scene representation.
The key idea is to use gravity as a universal physical prior: we first align the reconstruction to a unified Gravity-View frame to resolve gauge ambiguity, then recover object-centric rigid meshes with accurate 6-DoF placement, and finally remove duplicate object geometry from the background through conditional 3D point classification.
The resulting representation combines explicit rigid bodies with a decoupled background, enabling direct physics simulation while preserving visual realism.
Experiments on both simulated and real multi-view scenes show that \methodname improves object placement reliability, disentanglement quality, and rendering-simulation efficiency compared with retrieval-based baselines. Project page: \url{https://sunjiahaovo.github.io/garden/}

\end{abstract}

%%%%%%%%% BODY TEXT
\section{Introduction}
\label{sec:intro}

While recent RGB-only reconstruction systems~\cite{vggt,da3,pi3,nerf,ngp,3dgs} achieve impressive visual quality, they typically produce weakly structured ``3D photographs'' that struggle to support physically grounded downstream applications. These outputs are largely monolithic, coupling foreground rigid objects and background geometry into a single representation that hinders independent object-level manipulation. Furthermore, they are defined only up to an arbitrary global rotation without an explicit gravity direction. Without this physically grounded global frame, rigid-body anchoring and physics-aware interactions in environments like Embodied AI~\cite{duan2022survey} and simulations~\cite{mujoco,isaac} become fundamentally unstable.

\begin{figure}[t]
    \centering
    \includegraphics[width=0.85\linewidth]{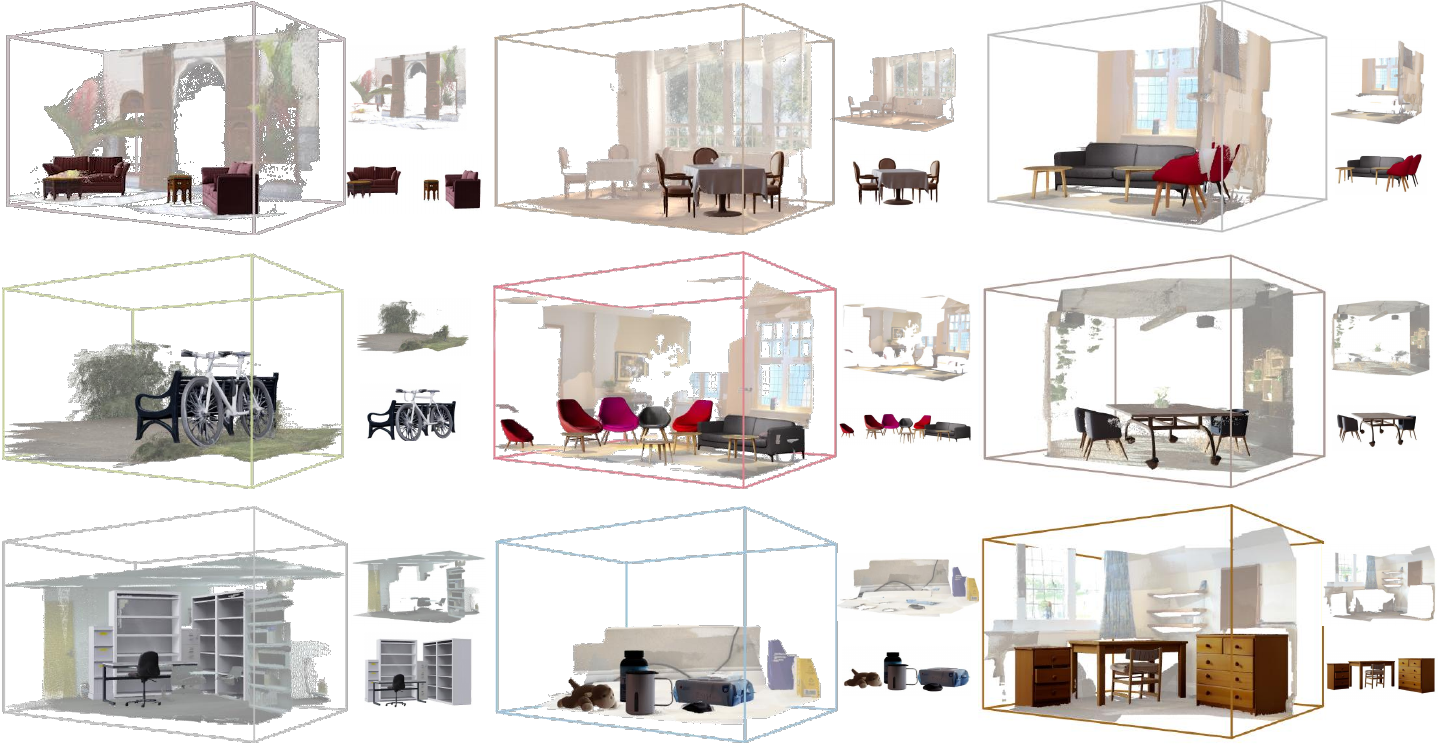}
    % \caption{From unstructured multi-view RGB inputs, our gravity-aligned and physically-grounded factorization produces a structured hybrid scene representation. Specifically, it reconstructs scene-specific rigid objects as independent meshes with consistent 6-DoF placement, explicitly aligning vertical orientations with the estimated gravity direction. Concurrently, removing duplicated object geometry ensures scene-faithful background reconstruction. Across diverse scenes, this decoupling preserves object authenticity and background realism, directly facilitating downstream interactive applications without CAD retrieval.}
    \caption{\textbf{GARDEN} factorizes unstructured RGB images into a structured, gravity-aligned representation. By decoupling scene-specific rigid objects from the background as meshes, the proposed method achieves high-fidelity reconstruction that enables interactive downstream simulation without the requirement of external CAD models.}
    \label{fig:teaser}
\end{figure}

To inject the necessary structure for interaction, existing pipelines often resort to CAD retrieval~\cite{im2cad, mask2cad, sparc, litereality}, which unfortunately introduces a severe trade-off between semantic structure and scene-specific realism. These systems typically extract layout bounding boxes and replace real objects with retrieved artist-designed 3D assets. However, this CAD substitution compromises \textit{instance-level geometric fidelity}, as pre-modeled assets rarely match the exact shape and fine-grained details of in-the-wild objects. Moreover, these pipelines are complex and brittle, accumulating errors across layout estimation, retrieval, and alignment stages. Finally, scaling this approach to diverse, long-tailed real-world objects is difficult, often requiring manual refinement.

In this work, our goal is to eliminate this compromise between structure and fidelity by proposing a \textbf{structured hybrid representation}. We seek a representation that is \textit{both} structurally disentangled for independent object manipulation \textit{and} strictly faithful to the scene-specific geometry captured in the real world. As visualized in \cref{fig:teaser}, our approach decouples foreground rigid objects—represented as explicit standalone meshes with accurate 6-DoF poses—from a clean background environment, represented as high-fidelity point clouds or 3DGS. This combination of physical independence and geometric authenticity directly facilitates stable downstream simulation.

To construct this representation without CAD retrieval, our core insight is that gravity provides a universal physical prior to resolve gauge ambiguity and anchor objects consistently. We introduce \methodname, an RGB-only pipeline that factorizes unstructured multi-view images into our structured hybrid representation. First, we resolve gauge ambiguity by extracting global camera tokens from a multi-view foundation model to regress the gravity direction, establishing a unified Gravity-View~\cite{gvhmr} frame where object placement is naturally constrained to maintain upright orientations. Next, guided by a lightweight target box that can be supplied by a user or generated by a vision-language model, we reconstruct scene-specific objects using 3D foundation models~\cite{sam3d} and refine their 6-DoF poses~\cite{foundationpose} within this gravity-aligned space. Finally, to eliminate object-background entanglement, we design a conditional 3D point classification network that identifies and removes redundant object geometry from the background, yielding a clean environment for hybrid rendering.

We evaluate \methodname against state-of-the-art scene generation and retrieval pipelines. By replacing heavy CAD retrieval and layout estimation with physically grounded factorization, our method substantially reduces processing time (e.g., from 4330s in LiteReality~\cite{litereality} to 560s), while improving geometric accuracy and rendering fidelity. In summary, our main contributions are:
\begin{itemize}
    \item We propose a novel paradigm of \textit{physically-grounded scene factorization} for RGB-only reconstruction, utilizing gravity as a physical prior to resolve gauge ambiguity and enable consistent rigid-body anchoring.
    \item We design a complete, CAD-free pipeline that extracts scene-specific explicit object meshes and accurately refines their 6-DoF poses within a unified Gravity-View coordinate frame.
    \item We introduce a conditional 3D point classification network to remove object redundancy from the background, enabling a \textit{structured hybrid representation} with strong rendering fidelity and computational efficiency.
\end{itemize}

\section{Related Work}
\label{sec:related}

\paragraph{Non Factorizable Scene Reconstruction}
The task of 3D surface reconstruction has been comprehensively investigated via both optimization-driven and learning-based paradigms~\cite{kinectfusion, poissonsurface, niessner2013real}. While recent advancements, including NeRF~\cite{nerf}, 3DGS~\cite{3dgs}, and subsequent models~\cite{mipnerf, zipnerf, ngp, mipsplatting}, produce exceptional quality in novel view synthesis, the primary focus of these techniques remains on visual rendering rather than the extraction of precise underlying geometry. To address the limitations in geometric accuracy, implicit representations based on SDF~\cite{neuralangelo, neus, yariv2021volume} have been developed to ensure geometric fidelity without compromising the quality of novel views. Furthermore, to bypass the extensive computational costs of per-scene optimization, feed-forward architectures~\cite{dust3r, mast3r, vggt, pi3, da3} have been proposed to deduce the global geometry of scenes directly from unposed input images. Despite the efficiency of these approaches, the standard practices model entire scenes as holistic continuous surfaces. Consequently, individual object representations often suffer from severe incompleteness under occlusions. Moreover, these frameworks leave a fundamental ambiguity in global orientation unresolved. Lacking an explicitly defined gravity direction or physical vertical axis, this gauge freedom entangles rigid objects with the background environment. Such monolithic structures present significant challenges for consistently anchoring, independently manipulating, or physically simulating individual entities within the reconstructed scene.

\paragraph{Factorizable Scene Reconstruction}
The creation of factorizable environments is a fundamental problem for advanced visual synthesis and physically-consistent scene composition. Early systems tackle the factorization of scenes through joint detection and completion~\cite{dahnert2021panoptic, runz2020frodo, hou2020revealnet}, procedural generation~\cite{ai2thor, habitat, behavior1k, ProcTHOR}, or the retrieval of CAD models~\cite{im2cad, mask2cad, sparc}. Recent pipelines, such as LiteReality~\cite{litereality} and other language-guided frameworks~\cite{holodeck, phone2proc, acdc}, facilitate the construction of interactive environments derived from real-world observations. HoloScene~\cite{xia2026holoscene} also targets simulation-ready interactive worlds from a single video with a scene-graph representation. However, many practical systems rely on proxy asset replacement. While this strategy supports basic scene arrangement, it frequently causes the reconstructed environments to drift from the exact geometry of the original objects. 
To better preserve object specificity, recent progress reframes reconstruction as conditional generation. Compositional pipelines~\cite{gen3dsr, midi, cast, sam3d} utilize diffusion priors and generative models to assemble coherent scenes from single images. Despite offering open-vocabulary flexibility, these generative techniques struggle to integrate stably with multi-view reconstructed environments. More fundamentally, isolating entities from standard neural representations often inherits the aforementioned global orientation ambiguity. Lacking an explicitly defined gravity direction, extracted objects lack physically consistent global anchoring, hindering reliable 3D manipulation or rearrangement.
To address these critical limitations, the proposed method reconstructs independent 3D shapes that are explicitly grounded within a physical coordinate system. Compared to prior techniques that may degrade across diverse scenes, our approach remains robust to casual capture conditions and successfully resolves the inherent gauge freedom. Consequently, this ensures that individual objects can be consistently anchored and independently manipulated for realistic scene simulation.

\section{Method}
\label{sec:method}

\begin{figure}[t]
    \centering
    \includegraphics[width=0.9\linewidth]{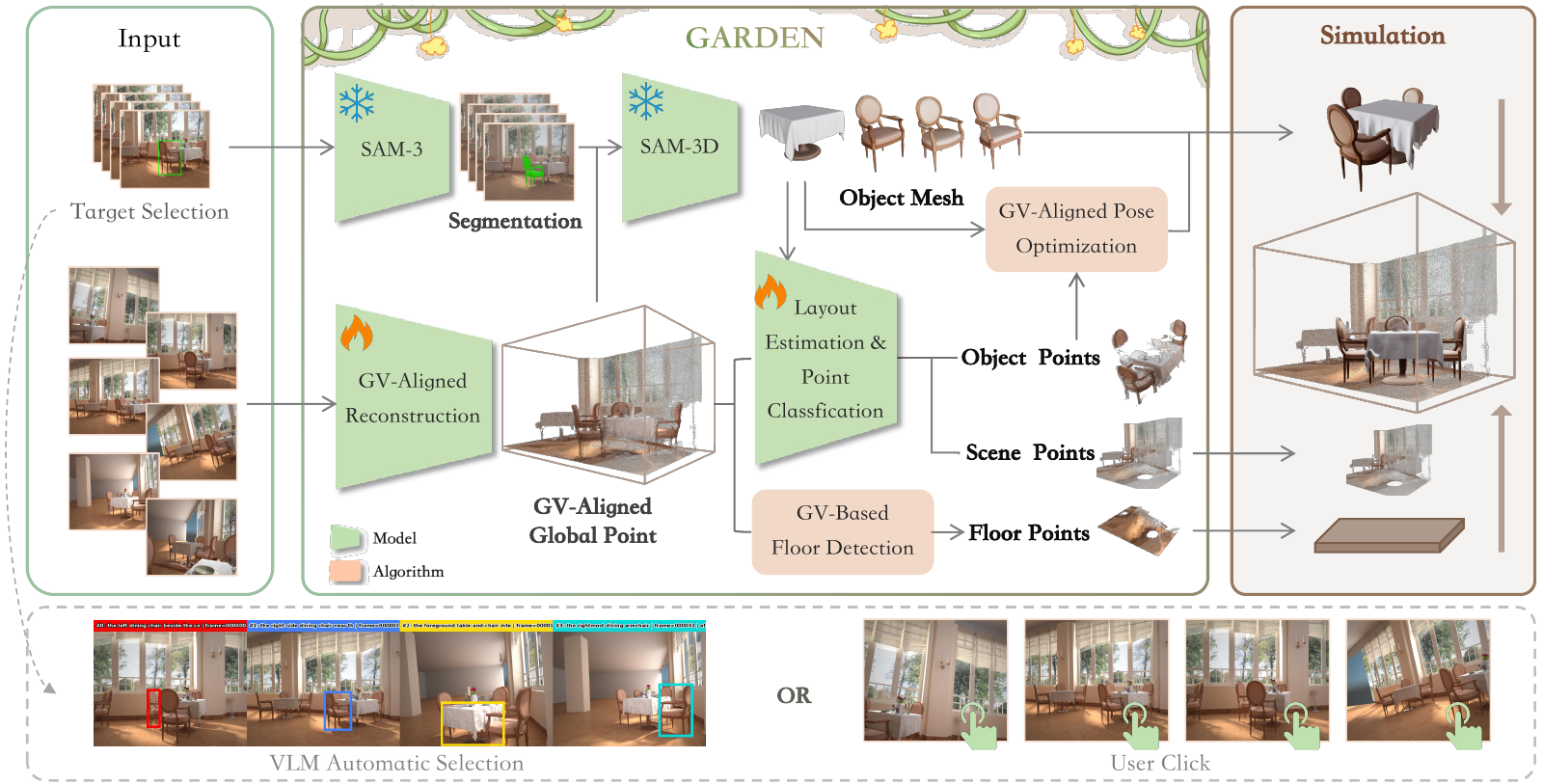}
    \caption{\textbf{Overview of \methodname. }
    Given multi-view RGB images, we first perform (1) \textit{multi-view reconstruction with gravity alignment} to establish a physically grounded coordinate system. Next, we conduct (2) \textit{target-driven object generation and layout estimation} to reconstruct standalone rigid objects from user- or VLM-provided boxes. This is followed by (3) \textit{point-conditional background disentanglement} to remove duplicate object geometry from the static scene. Finally, these decoupled components enable (4) \textit{unified physics simulation} with high-fidelity rendering.
    }
    \label{fig:pipeline}
\end{figure}

\methodname targets multi-view scene reconstruction for applications that require both physical interaction and high-fidelity rendering. 
Conventional multi-view reconstruction~\cite{colmap,vggt,da3} produces a single monolithic geometry without explicit object-level separation, and the reconstructed scene is defined up to an arbitrary rotation, and is therefore not physically grounded.
These properties limit structured scene decomposition and object-level physical interaction.

We formulate the problem as \textit{physically-grounded scene factorization} driven by gravity. 
As illustrated in \cref{fig:pipeline}, we first establish a unified Gravity-View~\cite{gvhmr} (GV) coordinate frame to resolve global orientation ambiguity, and then decompose the scene into independent rigid bodies and a static background.
Specifically, user-specified interactive objects are represented as standalone rigid meshes with accurate 6-DoF poses to enable simulation, while the background is represented by colored point clouds or 3D Gaussian Splatting (3DGS) to preserve visual fidelity. 
This hybrid representation supports both physical simulation (e.g., MuJoCo~\cite{mujoco}) and photorealistic rendering.
% Each stage is described in detail below.

\subsection{Multi-View Reconstruction with Gravity-View Alignment}

Unlike methods~\cite{litereality} that rely on depth or IMU sensors, our approach generates scene reconstructions solely from multi-view RGB images. 
This is facilitated by recent advancements in multi-view foundation models (e.g., VGGT~\cite{vggt}, Pi3~\cite{pi3}, and DepthAnything-3~\cite{da3}), which enable the recovery of camera poses and scene structures from pure visual input. 
However, a significant challenge remains: the global coordinate systems of these models are implicitly determined by their training priors. 
For instance, VGGT and DepthAnything-3 typically anchor the scene to a reference camera frame, while Pi3 employs a permutation-equivariant approach that yields an indeterminate global orientation. 
Consequently, performing direct scene factorization on such reconstructions is hindered by global orientation ambiguities, and the recovered entities lack the gravity alignment required for downstream physical simulations.
To address this, inspired by GVHMR~\cite{gvhmr}, we propose the Gravity-View Align Module to align the reconstructed coordinate system with the physical direction of gravity.

\begin{figure}[t]
    \centering
    \includegraphics[width=0.5\linewidth]{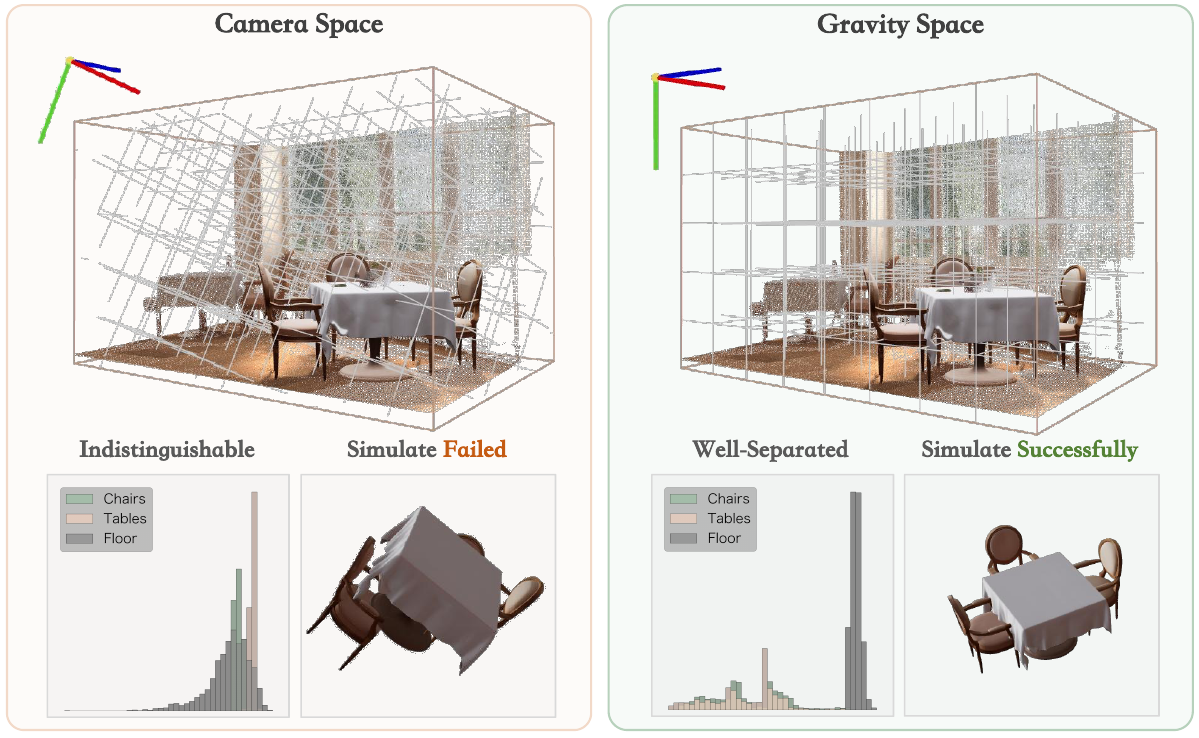}
    \caption{\textbf{Motivation for Gravity-View (GV) alignment.} Point clouds reconstructed in the camera frame are globally tilted, whereas GV alignment restores an upright geometry consistent with physical gravity. This correction improves structural separability, as evidenced by the histogram in this figure: floor and object points, heavily entangled in the camera frame, become more distinguishable in GV space and therefore easier to factorize. In downstream physics simulation, GV-aligned scenes remain stable under gravity, while camera-frame reconstructions are gravity-inconsistent and tend to produce unstable object poses (e.g., tilting or falling).}
    \label{fig:gv}
\end{figure}

We first construct the GV coordinate system by aligning the reference frame with the gravity direction. 
Taking the standard OpenCV camera convention as a reference: 
(1) within the camera coordinate system, we define the gravity direction as $\mathbf{g}$; 
(2) the y-axis of the GV frame is aligned with the gravity direction, such that $\mathbf{y} = \mathbf{g}$; 
(3) given the view direction in the camera frame $\mathbf{v} = [0, 0, 1]^\top$, we derive the x-axis of the GV frame as $\mathbf{x} = \mathbf{y} \times \mathbf{v}$; 
(4) finally, the z-axis is determined via the right-hand rule as $\mathbf{z} = \mathbf{x} \times \mathbf{y}$. 
Through this construction, the GV coordinate system is established, allowing for the direct derivation of the rotation matrix $\mathbf{R}_{c2gv}$ that transforms the original camera coordinates to the GV frame.
As illustrated in \cref{fig:gv}, this alignment is not merely a coordinate normalization step: it improves geometric separability between support structures and objects, and is critical for stable downstream physical simulation.

% Having established the Gravity-View coordinate system, the remaining task is to accurately predict this transformation from the visual input. 
% Since the multi-view foundation model~\cite{da3} used for reconstruction is trained on massive datasets, it inherently encodes strong 3D structural priors. 
% Critical geometric indicators of gravity—such as vertical walls and orthogonal corners—are deeply embedded in its feature representations. 
% Therefore, rather than training an alignment network from scratch, we repurpose these robust multi-view features to drive our Gravity-View decoding.

To predict the transformation to the GV frame, we leverage the multi-view foundation model Depth-Anything-3~\cite{da3} for its robust scene representation capabilities.
Specifically, we extract the global camera tokens $\mathbf{F} \in \mathbb{R}^{N \times C}$ from the backbone across the $N$ input images. 
The base model dynamically identifies a reference view $c_{ref}$, from which we select its corresponding token $\mathbf{f}_{ref} \in \mathbb{R}^{1 \times C}$ to serve as the query. To effectively aggregate the global multi-view context, we employ a Context Transformer Decoder that takes $\mathbf{f}_{ref}$ as the query and the complete token sequence $\mathbf{F}$ as keys and values. This cross-attention mechanism allows the network to gather gravity-aware geometric clues from all views. The output feature is then processed by an MLP head to regress a continuous 6D rotation representation, ensuring stable optimization. This 6D representation is finally mapped into a $3 \times 3$ rotation matrix, denoting the rotation from the reference view to the GV frame, $\hat{\mathbf{R}}_{c_{ref} \to gv}$.
% TODO: network arch in supp

For supervision, we derive the ground-truth alignment matrix $\mathbf{R}_{c \to gv}^{gt}$ by leveraging artist-designed simulation datasets, where the global coordinate systems are inherently gravity-aligned. Assuming the gravity direction in the world coordinate system is defined as $\mathbf{g}_{w} = [0, 0, -1]^\top$, given the ground-truth camera extrinsics $\mathbf{R}_{w \to c}$, we first project the world gravity vector into the camera coordinate system to obtain the Gravity-View y-axis: $\mathbf{y} = \mathbf{R}_{w \to c} \mathbf{g}_{w}$. 
Following the aforementioned GV formulation, given the camera view direction $\mathbf{v} = [0, 0, 1]^\top$, the x-axis is computed as $\mathbf{x} = \frac{\mathbf{y} \times \mathbf{v}}{\| \mathbf{y} \times \mathbf{v} \|}$. 
To ensure robustness against collinear singularities (i.e., when the camera looks directly along the gravity axis), we fall back to the camera's original x-axis if the cross-product norm approaches zero. 
The z-axis is subsequently derived via $\mathbf{z} = \mathbf{x} \times \mathbf{y}$. 
Finally, the ground-truth rotation matrix is formulated as $\mathbf{R}_{c \to gv}^{gt} = [\mathbf{x}, \mathbf{y}, \mathbf{z}]^\top$. 

During training, the module is optimized by applying an $L_1$ loss directly between the predicted rotation matrix of the reference view and its corresponding ground truth:
\begin{equation}
    \mathcal{L}_{rot} = \| \hat{\mathbf{R}}_{c_{ref} \to gv} - \mathbf{R}_{c_{ref} \to gv}^{gt} \|_1
\end{equation}
Once $\hat{\mathbf{R}}_{c_{ref} \to gv}$ is accurately predicted, the entire reconstructed scene can be robustly transformed into the gravity-aligned coordinate system, facilitating subsequent physical simulations.

\subsection{Target-Driven Object Generation and Layout Estimation}

We decouple task-relevant objects through a lightweight target-selection interface. In the simplest mode, the target is specified by a single 2D box on a reference view; alternatively, the same box can be generated by a vision-language model or open-vocabulary grounding module without changing any downstream stage.

Given the bounding box, we first employ SAM-3~\cite{sam3} to extract a precise 2D mask. This mask is then processed by SAM-3D~\cite{sam3d} to lift 2D semantics into 3D space.
We choose SAM-3D specifically for its amodal reconstruction capability, which recovers complete geometry even under partial occlusion.
While SAM-3D inherently predicts an initial spatial layout alongside the generated mesh, our empirical evaluations reveal that this native pose estimation is often insufficiently accurate for strict physical simulations. 

\begin{figure}[t]
    \centering
    \includegraphics[width=0.8\linewidth]{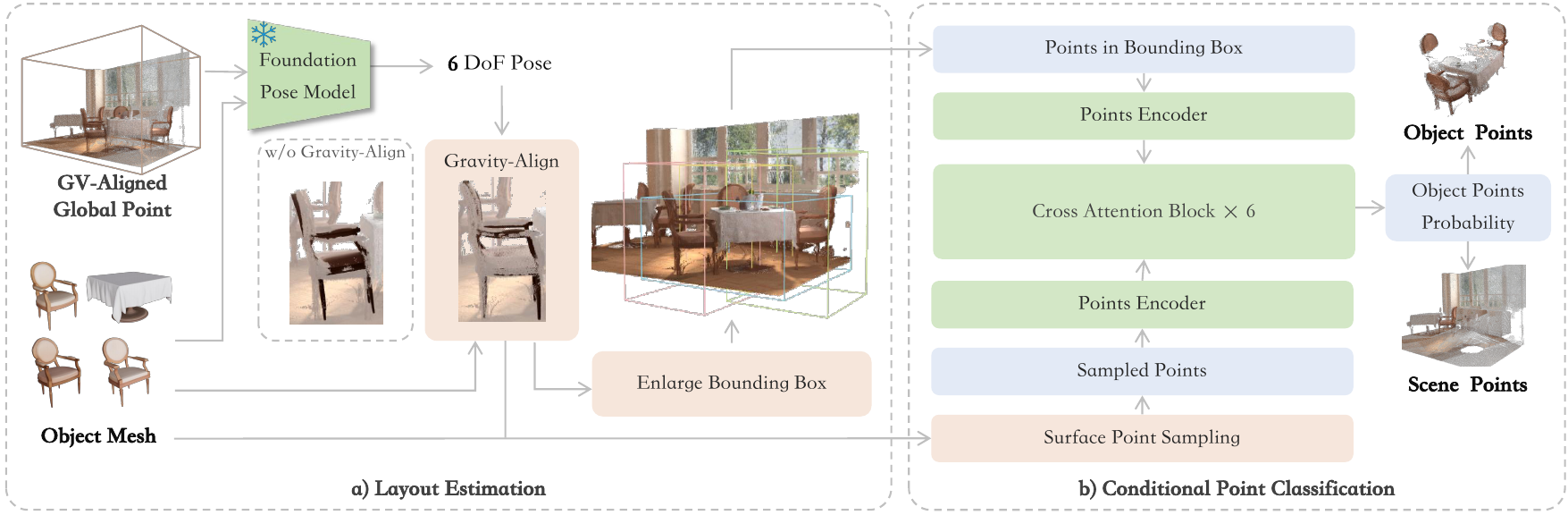}
    \caption{\textbf{Layout Estimation and background disentanglement in \methodname.} \textbf{(a) Layout Estimation: }Given a generated object mesh and the Gravity-View (GV) aligned global reconstruction, we first use FoundationPose~\cite{foundationpose} to estimate 6-DoF pose parameters and physically align with gravity direction. The \textit{w/o Gravity Align} inset illustrates that omitting this alignment causes physically inconsistent placement. We then build a relaxed 3D region by gravity-aligned box expansion. \textbf{(b) Conditional Point Classification: }Cropped scene points within the box and mesh-derived condition points are fed into our conditional point classifier. The network predicts point-level labels to separate target-object from static scene points, removing duplicated geometry while preserving high-fidelity background.}
    \label{fig:method}
\end{figure}

To refine the spatial layout, we integrate FoundationPose~\cite{foundationpose} as a dedicated module to compute the 6-DoF pose of the reconstructed mesh (\cref{fig:method}(a)). By leveraging the gravity direction through our GV Alignment Module, we transform the placement into a physically grounded problem. Since SAM-3D meshes are inherently axis-aligned, a deterministic rotation first aligns their vertical axis with gravity. FoundationPose then only needs to estimate the translation and yaw angle, effectively reducing the rotational search space. Finally, the locally aligned mesh is transformed into the global Gravity-View space via $\mathbf{R}_{c_{ref} \to gv}$ for downstream simulation.

\subsection{Point-Conditional Background Disentanglement}
\label{sec:dp}

To complete the physically-grounded factorization motivated in \cref{sec:intro}, placing an explicit object mesh is not sufficient by itself: we must also remove the duplicated object geometry that still remains in the original monolithic scene representation. Otherwise, the same object is represented twice, re-introducing foreground-background entanglement and causing visual and physical inconsistencies in downstream simulation. A naive 3D box crop is too coarse, because it often removes nearby background structures (e.g., supporting floors or occluding walls). Cross-view consistent 3D segmentation remains brittle, particularly in indoor scenes with repeated instances (e.g., similar chairs). Long-range association errors often cause incomplete or incorrect object removal.

% To preserve scene-specific fidelity while achieving structural disentanglement, we cast this step as a conditional 3D point classification problem as shown in \cref{fig:method}(b). We design a Transformer-based Point Classification Network that takes two inputs: scene points cropped with a relaxed box, and condition points sampled from the generated object mesh. Both point sets are embedded into a shared feature space using coordinate and type-specific embeddings, and processed by a multi-layer Transformer with self- and cross-attention to model fine-grained scene-object correspondences. The network predicts a binary probability for each scene point, indicating whether it belongs to the target object or the static background, and is trained with a standard Binary Cross-Entropy (BCE) loss.

To maintain scene-specific fidelity while achieving structural disentanglement, we formulate this task as a conditional 3D point classification problem, as illustrated in \cref{fig:method}(b). We developed a Transformer-based point classification network that processes two distinct inputs: scene points cropped via a relaxed bounding box and condition points sampled from the generated object mesh. Both point sets are mapped into a shared feature space using coordinate and type-specific embeddings. A multi-layer Transformer architecture, utilizing self- and cross-attention mechanisms, then models fine-grained correspondences between the scene and the object. The network predicts a binary probability for each scene point to determine its membership to either the target object or the static background, guided by a standard Binary Cross-Entropy (BCE) loss during training.

To avoid manual point-level annotations on real scans, we leverage InternScenes~\cite{internscenes}, a large-scale simulation dataset providing paired scene and object meshes for precise label construction. We bridge the domain gap between clean simulations and artifact-prone real reconstructions by introducing artifact-oriented 3D augmentations during training. These techniques, including multi-layer ghosting, thin-part deformation, virtual multi-view scanning, and jittering of mesh pose and scale, simulate systematic reconstruction errors. Training with these realistic perturbations enables robust simulation-to-reality transfer, ensuring reliable point-level disentanglement and cleaner object removal in real scenes while preserving background geometry. Detailed augmentation specifications are provided in the \textit{Supplementary Material}.

% \begin{figure}
%     \centering
%     \includegraphics[width=0.5\linewidth]{fig/dp pcl.pdf}
%     \caption{DP pcl}
%     \label{fig:dp}
% \end{figure}

\subsection{Unified Physics Simulation}

The final stage of our pipeline bridges the gap between static 3D reconstruction and interactive physical simulation. By leveraging the decoupled scene representation and the established Gravity-View coordinate system, we seamlessly construct a unified environment within a physics engine (e.g., MuJoCo~\cite{mujoco}). 

To physically ground the interactions, we first establish the static environment. Thanks to the gravity-aligned multi-view consistency, we identify the supporting floor plane by extracting the Top-K coordinate peaks along the gravity vector. This floor mesh is instantiated within the simulator as a static collision body. Concurrently, the user-specified objects—which have been reconstructed as complete local meshes and accurately localized via our layout estimation module—are imported as independent, dynamic rigid bodies. 

A key advantage of our pipeline is the inherent physical consistency provided by the Gravity-View frame. Since all scene components (points, local meshes, and layout poses) share this unified coordinate system, we can directly align the simulator's gravity vector with the canonical vertical axis (e.g., the $+y$-axis). 

Furthermore, we adopt a dual-representation strategy for the final environment: while the extracted floor and object meshes govern the rigid-body collision dynamics, the remaining cleaned background representation (such as the colored point cloud or 3DGS) is rendered as a high-fidelity visual backdrop. This design ensures that the generated environment not only strictly obeys physical laws for downstream embodied AI interactions but also preserves the photorealistic visual context of the original scene.

\section{Experiments}
\label{sec:exp}

Following the LiteReality~\cite{litereality} protocol, we evaluate \methodname from three complementary perspectives: object-centric perceptual quality, holistic scene quality, and end-to-end computational efficiency. We further conduct two additional diagnostic experiments on gravity estimation accuracy and NVS-based factorization ablation.

\subsection{Experimental Setup}

\subsubsection{Benchmarks and Evaluation Protocols}

We use three LiteReality evaluation settings: object-centric assessment, holistic scene assessment, and computational efficiency. In addition, we evaluate gravity estimation accuracy and conduct an NVS-based ablation to isolate the effectiveness of our factorization design.

\textit{Object-Centric Assessment: }We adopt the \textit{object-centric material recovery} benchmark from LiteReality~\cite{litereality}. In the standard baseline protocol, methods are evaluated in a static setting using accurate cropped object bounding boxes. In contrast, we evaluate our method under a simulation-ready protocol: reconstructed assets are imported into MuJoCo~\cite{mujoco}, simulated for 10 seconds under gravity, and then rendered for metric computation using the user-provided query bounding box as ROI. This setting explicitly penalizes both appearance errors and physically induced pose/layout errors.

\textit{Holistic Scene Assessment: }
We adopt the \textit{graphics-ready reconstruction} benchmark~\cite{litereality}, which measures similarity between input RGB frames and rendered reconstructed scenes. Different from static evaluation, our scene-level results are also measured after the same 10-second MuJoCo~\cite{mujoco} simulation. Therefore, floating, intersection, or gravity-inconsistent placement errors directly reduce the final perceptual scores.

\textit{Gravity Estimation Assessment: }
We evaluate the predicted gravity direction by angular error against ground truth on held-out Hypersim~\cite{hypersim} and TartanAir~\cite{tartanair} scenes. We report mean error, 90th percentile error, and the failure rate above $10^\circ$.

\textit{NVS Ablation Assessment: }
To validate our design choices, we perform NVS-based ablation on 5 Hypersim~\cite{hypersim} scenes. We skip full-scene evaluation for point-cloud geometry. Given that NVS benchmarks provide limited viewpoints, under-observed regions degrade metrics and reduce method discriminability. We instead compute distances within an object-centric local region (including immediate support) where methods are comparably observable. This ablation jointly evaluates rendering quality and local reconstruction, revealing factorization effects beyond direct 6-DoF placement.

\textit{Computational Efficiency: }
We compare inference latency with LiteReality~\cite{litereality} on a workstation with an Intel Xeon w7-3445 CPU and an NVIDIA RTX A6000 GPU. We exclude evaluation-specific rendering overhead.

\subsubsection{Baselines and Variants}

\textit{Baselines: }
For object-centric assessment, we compare against PhotoShape~\cite{photoshape} and MIR~\cite{mir}, together with LiteReality~\cite{litereality}-related combinations and variants, including Visual+Language Search (Sem\&Vis), MIR + Albedo-Only Optimization (AO), and the full LiteReality system. For holistic scene assessment, we compare against Phone2Proc~\cite{phone2proc} and ACDC~\cite{acdc}, as well as ACDC + MIR, ACDC + Sem\&Vis, and LiteReality.
For gravity estimation, we compare against Plane-RANSAC, normal clustering, COLMAP's Manhattan alignment~\cite{colmap}, GeoCalib~\cite{veicht2024geocalib}, and GeoCalib+RANSAC.
\textit{Variants: }
For NVS ablation, we compare \textbf{Ours} with representative variants: \textbf{DA3}, which directly uses the reconstruction backbone---DepthAnything3~\cite{da3} output; \textbf{w/ GV, full FP pose w/o sim}, which places generated objects via FoundationPose~\cite{foundationpose} without the complete physically-grounded integration; and two \textbf{w/o GV, full FP pose} variants that remove the learned Gravity-View alignment.

\subsubsection{Implementation Details}

Our pipeline is implemented in PyTorch, and all trainable modules are optimized using AdamW. Training is conducted on 32 NVIDIA H20 GPUs.
\textit{Gravity-View Align Module:} We freeze the pre-trained DepthAnything3~\cite{da3} backbone and train the alignment module for 10k steps on TartanAir~\cite{tartanair}, Hypersim~\cite{hypersim}, and vKitti~\cite{vkitti}. The peak learning rate is $5 \times 10^{-6}$ with a 1k-step linear warm-up. The batch size is dynamically adjusted to keep an approximately constant token count per step.
\textit{Conditional Background Point Classification Module:} We train the classification network for 2k steps on preprocessed InternScenes~\cite{internscenes} with a $5 \times 10^{-4}$ base learning rate and batch size 48.
\textit{Inference:} Input multi-view images are resized to a maximum side length of 504 (following DepthAnything3~\cite{da3}) to recover camera poses, global point clouds, and 3DGS. For object generation, we cascade SAM-3~\cite{sam3} (2D mask extraction) and SAM-3D~\cite{sam3d} (amodal mesh reconstruction). FoundationPose~\cite{foundationpose} refines the layout with 5 iterations. In background disentanglement, we remove object-associated points using a 0.5 probability threshold. For simulation-backed evaluations, we use MuJoCo~\cite{mujoco} with $0.001$s timestep, 50-iteration Newton solver per step, and a pyramidal friction cone. The gravity vector $\mathbf{g}_{sim} = [0, 9.81, 0]^\top$ is set to align with our Gravity-View coordinate system.

\subsection{Results}

% \subsubsection{Main Comparison with Baselines}

\begin{table}[t]
    \centering
    \caption{\textbf{Object-centric material recovery comparison.} We follow LiteReality's~\cite{litereality} object benchmark and report RMSE/SSIM/LPIPS. Baselines are evaluated with the standard static cropped-bbox protocol. Our main results use the stricter post-MuJoCo~\cite{mujoco} stabilization protocol, and static GARDEN results are included for protocol comparison. Best results are highlighted as \colorbox[HTML]{ACD6B7}{\textbf{first}}, 
    \colorbox[HTML]{CFDDA8}{second}, 
    \colorbox[HTML]{E8E6B0}{third}. } 
    \label{tab:object}
    \begin{tabular}{l|ccc}
        \toprule
        Methods & RMSE $\downarrow$ & SSIM $\uparrow$ & LPIPS $\downarrow$ \\
        \midrule
        MIR~\cite{mir} & 0.2377 & 0.3981 & 0.6111 \\
        PhotoShape~\cite{photoshape} & 0.3225 & 0.2371 & 0.6558 \\
        MIR~\cite{mir} + AO~\cite{litereality} & \colorbox[HTML]{E8E6B0}{0.2156} & \colorbox[HTML]{E8E6B0}{0.4203} & 0.5899 \\
        Sem\&Vis~\cite{litereality} & 0.2835 & 0.3758 & 0.6362 \\
        LiteReality~\cite{litereality} & 0.2163 & \colorbox[HTML]{E8E6B0}{0.4353} & 0.5854 \\
        \textbf{Ours (3DGS, post-sim)} & \colorbox[HTML]{E8E6B0}{0.1880} & 0.4181 & 0.5035 \\
        \textbf{Ours (point cloud, post-sim)} & 0.1887 & 0.4240 & \colorbox[HTML]{E8E6B0}{0.4444} \\
        \midrule
        \textbf{Ours (3DGS, static)} & \colorbox[HTML]{CFDDA8}{0.1796} & \colorbox[HTML]{CFDDA8}{0.4494} & \colorbox[HTML]{CFDDA8}{0.4182} \\
        \textbf{Ours (point cloud, static)} & \colorbox[HTML]{ACD6B7}{\textbf{0.1736}} & \colorbox[HTML]{ACD6B7}{\textbf{0.4656}} & \colorbox[HTML]{ACD6B7}{\textbf{0.3680}} \\
        % \midrule
        % da3(gs) & 0.1380 & 0.4983 & 0.4448 \\
        % da3(point cloud) & 0.1370 & 0.5232 & 0.3721 \\
        % \midrule
        % full fp pose w sim(gs) & 0.1981 & 0.3982 & 0.5547 \\
        % full fp pose w sim(point cloud) & 0.2006 & 0.4005 & 0.4665 \\
        % \midrule
        % full fp pose wo sim (gs) & 0.1826 & 0.4341 & 0.4262 \\
        % full fp pose wo sim (point cloud) & 0.1879 & 0.4369 & 0.4144 \\
        \bottomrule
    \end{tabular}
\end{table}
\begin{table}[t]
    \centering
    \caption{\textbf{Holistic graphics-ready scene comparison.} We report full-scene RMSE/SSIM/LPIPS between rendered reconstructions and input frames. Our main rendering is measured after physical simulation, directly testing gravity-consistent placement and global scene stability; static GARDEN results are included for protocol comparison. Best results are highlighted as \colorbox[HTML]{ACD6B7}{\textbf{first}}, 
    \colorbox[HTML]{CFDDA8}{second}, 
    \colorbox[HTML]{E8E6B0}{third}. }
    \label{tab:scene}
    \begin{tabular}{l|ccc}
        \toprule
        Methods & RMSE $\downarrow$ & SSIM $\uparrow$ & LPIPS $\downarrow$ \\
        \midrule
        Phone2Proc~\cite{phone2proc}        & 0.3604 & 0.5512 & 0.7338 \\
        ACDC~\cite{acdc}    & 0.3653 & 0.5531 & 0.7364 \\
        ACDC~\cite{acdc} + Sem\&Vis~\cite{litereality}     & 0.3226 & 0.5425 & 0.6717 \\
        ACDC~\cite{acdc} + MIR~\cite{mir}    & 0.3046 & 0.5492 & 0.6648 \\
        LiteReality~\cite{litereality}    & 0.2664 & \colorbox[HTML]{CFDDA8}{0.5818} & 0.6522 \\
        \textbf{Ours (3DGS, post-sim)} & 0.1670 & 0.5573 & 0.4709 \\
        \textbf{Ours (point cloud, post-sim)} & \colorbox[HTML]{E8E6B0}{0.1616} & \colorbox[HTML]{E8E6B0}{0.5735} & \colorbox[HTML]{CFDDA8}{0.4081} \\
        \midrule
        \textbf{Ours (3DGS, static)} & \colorbox[HTML]{CFDDA8}{0.1593} & 0.5718 & \colorbox[HTML]{E8E6B0}{0.4379} \\
        \textbf{Ours (point cloud, static)} & \colorbox[HTML]{ACD6B7}{\textbf{0.1511}} & \colorbox[HTML]{ACD6B7}{\textbf{0.5916}} & \colorbox[HTML]{ACD6B7}{\textbf{0.3770}} \\
        % \midrule
        % da3(gs) & 0.1353 & 0.5953 & 0.4315 \\
        % da3(point cloud) & 0.1424 & 0.6073 & 0.3788 \\
        % \midrule
        % full fp pose w sim(gs) & 0.1770 & 0.5484 &  0.4977 \\
        % full fp pose w sim (point cloud) & 0.1941 & 0.5402 & 0.4660 \\
        % \midrule
        % full fp pose wo sim (gs) & 0.1680 & 0.5639 &  0.4544 \\
        % full fp pose wo sim (point cloud) & 0.1778 & 0.5676 & 0.4248 \\
        \bottomrule
    \end{tabular}
\end{table}

\paragraph{Object-Centric Assessment.}
\cref{tab:object} shows that our method consistently improves over LiteReality~\cite{litereality} in RMSE and LPIPS, while keeping SSIM close. Specifically, under the stricter post-simulation protocol, Ours (3DGS) reduces RMSE from 0.2163 to 0.1880, and Ours (point cloud) reduces LPIPS from 0.5854 to 0.4444. The static GARDEN rows further improve these scores, confirming that the gains are not an artifact of post-simulation evaluation. Although LiteReality reports strong SSIM under its static protocol, the gap is small. We attribute this primarily to two factors: (1) unlike LiteReality, our pipeline does not include a dedicated surface-texture prediction stage; and (2) our post-simulation evaluation introduces minor physically induced pose perturbations, to which SSIM is sensitive. Overall, the stronger RMSE/LPIPS results indicate better geometry-aware and perceptual fidelity under a stricter simulation-ready protocol.

\paragraph{Holistic Scene Assessment.}
\cref{tab:scene} demonstrates consistent scene-level gains over prior methods. Compared with LiteReality~\cite{litereality}, Ours (point cloud, post-sim) improves RMSE from 0.2664 to 0.1616 and LPIPS from 0.6522 to 0.4081, while maintaining comparable SSIM (0.5735 vs. 0.5818). The static rows again show stronger scores than the post-simulation rows, indicating that the post-simulation protocol is a conservative stress test rather than a source of artificial gains. This trend is consistent with the object-level findings: even without a dedicated texture prediction stage, our physically-grounded factorization provides stronger global scene fidelity and perceptual quality.

\subsubsection{Gravity Estimation Accuracy}

\cref{tab:gravity_estimation} evaluates GV prediction accuracy against classical and learned gravity-estimation baselines. The proposed GV predictor achieves the lowest mean and P90 angular errors on both datasets, with $0.0\%$ Fail@10. Plane-RANSAC and normal clustering can work when clean dominant planes are visible, but they are brittle under noisy or incomplete reconstructed geometry. GeoCalib~\cite{veicht2024geocalib} and COLMAP's Manhattan alignment~\cite{colmap} are stronger alternatives, yet they remain less accurate than our multi-view GV predictor.

\begin{table}[t]
    \centering
    \caption{\textbf{Gravity estimation accuracy.} Angular error (degrees) on held-out Hypersim~\cite{hypersim} and TartanAir~\cite{tartanair} scenes.}
    \label{tab:gravity_estimation}
    \setlength{\tabcolsep}{3pt}
    \begin{tabular}{l|ccc|ccc}
        \toprule
        \multirow{2}{*}{Method} &
        \multicolumn{3}{c|}{Hypersim} &
        \multicolumn{3}{c}{TartanAir} \\
        & Mean $\downarrow$ & P90 $\downarrow$ & Fail@10 $\downarrow$
        & Mean $\downarrow$ & P90 $\downarrow$ & Fail@10 $\downarrow$ \\
        \midrule
        Plane-RANSAC & 24.87 & 89.68 & 26.7 & 19.26 & 83.22 & 34.4 \\
        Normal clustering & 30.98 & 89.77 & 33.3 & 21.33 & 88.70 & 37.5 \\
        COLMAP's Manhattan & 2.62 & 6.38 & \textbf{0.0} & 8.37 & 15.91 & 12.5 \\
        GeoCalib~\cite{veicht2024geocalib} & 7.59 & 4.79 & 6.7 & 3.01 & 6.50 & 9.4 \\
        GeoCalib+RANSAC & 1.71 & 3.79 & \textbf{0.0} & 2.69 & 5.91 & 9.4 \\
        \textbf{Ours GV} & \textbf{1.40} & \textbf{1.90} & \textbf{0.0} & \textbf{1.56} & \textbf{3.48} & \textbf{0.0} \\
        \bottomrule
    \end{tabular}
\end{table}

\subsubsection{Ablation on Factorization and Placement}

\cref{tab:scene_nvs} validates the effectiveness of our factorization design. Compared with DA3(point cloud), Ours(point cloud) achieves better completeness (Comp., 0.0387 vs. 0.0516), while improving RMSE and SSIM in rendering. Although DA3 obtains lower Acc. and higher N.C., its weaker completeness suggests that it tends to recover a more limited subset of target-object-region geometry. Our method reconstructs more complete local object-centric geometry while preserving competitive rendering quality.
Compared with w/ GV, full FP pose w/o sim(point cloud), Ours(point cloud) is better across all reported metrics. Removing GV from the same full-FP-pose setting causes substantial degradation both without simulation and after simulation, confirming that GV improves pose initialization, background disentanglement, and downstream physical stability rather than serving as a cosmetic coordinate transform.

\begin{table}[t]
    \centering
    \caption{\textbf{NVS ablation on Hypersim~\cite{hypersim}.} We compare full \methodname with DA3 and full-FP-pose variants with or without GV using rendering metrics (RMSE/SSIM/LPIPS) and point-cloud geometry metrics (Acc./Comp./N.C.). Point-cloud metrics are computed in an object-centric local region.}
    \label{tab:scene_nvs}
    \begin{tabular}{l|ccc|ccc}
        \toprule
        Methods & RMSE $\downarrow$ & SSIM $\uparrow$ & LPIPS $\downarrow$ & Comp. $\downarrow$ & Acc. $\downarrow$ & N.C. $\uparrow$\\
        \midrule
        \rowcolor{gray!20}\multicolumn{7}{l}{\textit{3DGS Background}}\\
        DA3 & 0.3068 & 0.3141 & 0.6623 & - & - & -\\
        w/o GV, full FP pose w/o sim & 0.3134 & 0.3000 & 0.6617 & - & - & -\\
        w/o GV, full FP pose w/ sim & 0.3171 & 0.2885 & 0.6725 & - & - & -\\
        w/ GV, full FP pose w/o sim & 0.3077 & 0.3154 &  0.6460 & - & - & -\\
        \textbf{Ours} & \textbf{0.2840} &  \textbf{0.3809} & \textbf{0.5575} & - & - & -\\
        % \midrule
        % full fp pose w sim (gs) & 0.3081 & 0.3128 &  0.6431 & - & - & -\\
        % full fp pose w sim (point cloud) & 0.2749 & 0.3567 & 0.5344 & 0.0517 & 0.1653 & 0.7391\\
        \midrule
        \rowcolor{gray!20}\multicolumn{7}{l}{\textit{Point Cloud Background}}\\
        DA3 & 0.2788 & 0.3482 & 0.5311 & 0.0516 & \textbf{0.0082} & \textbf{0.9129}\\
        w/o GV, full FP pose w/o sim & 0.2961 & 0.3211 & 0.5917 & 0.0447 & 0.0726 & 0.7320\\
        w/o GV, full FP pose w/ sim & 0.3002 & 0.3096 & 0.5929 & 0.0743 & 0.3132 & 0.7078\\
        w/ GV, full FP pose w/o sim & 0.2753 & 0.3562 & 0.5351 & 0.0470 & 0.0674 & 0.7275\\
        \textbf{Ours} & \textbf{0.2751} & \textbf{0.3565} & \textbf{0.5338} & \textbf{0.0387} & 0.0658 & 0.7497\\
        \bottomrule
    \end{tabular}
\end{table}

The 3DGS background directly uses DA3's GS head; its lower NVS quality in this ablation is mainly caused by low-opacity Gaussian primitives, while the fixed background still isolates the relative effect of our factorization components.

\subsubsection{Qualitative Results}

We provide qualitative results under the same physically grounded protocol used throughout this paper: all visualizations are captured after a 10-second MuJoCo~\cite{mujoco} simulation. This setting evaluates not only appearance quality but also gravity-consistent placement and structural stability.

\begin{figure}[t]
    \centering
    \includegraphics[width=0.45\linewidth]{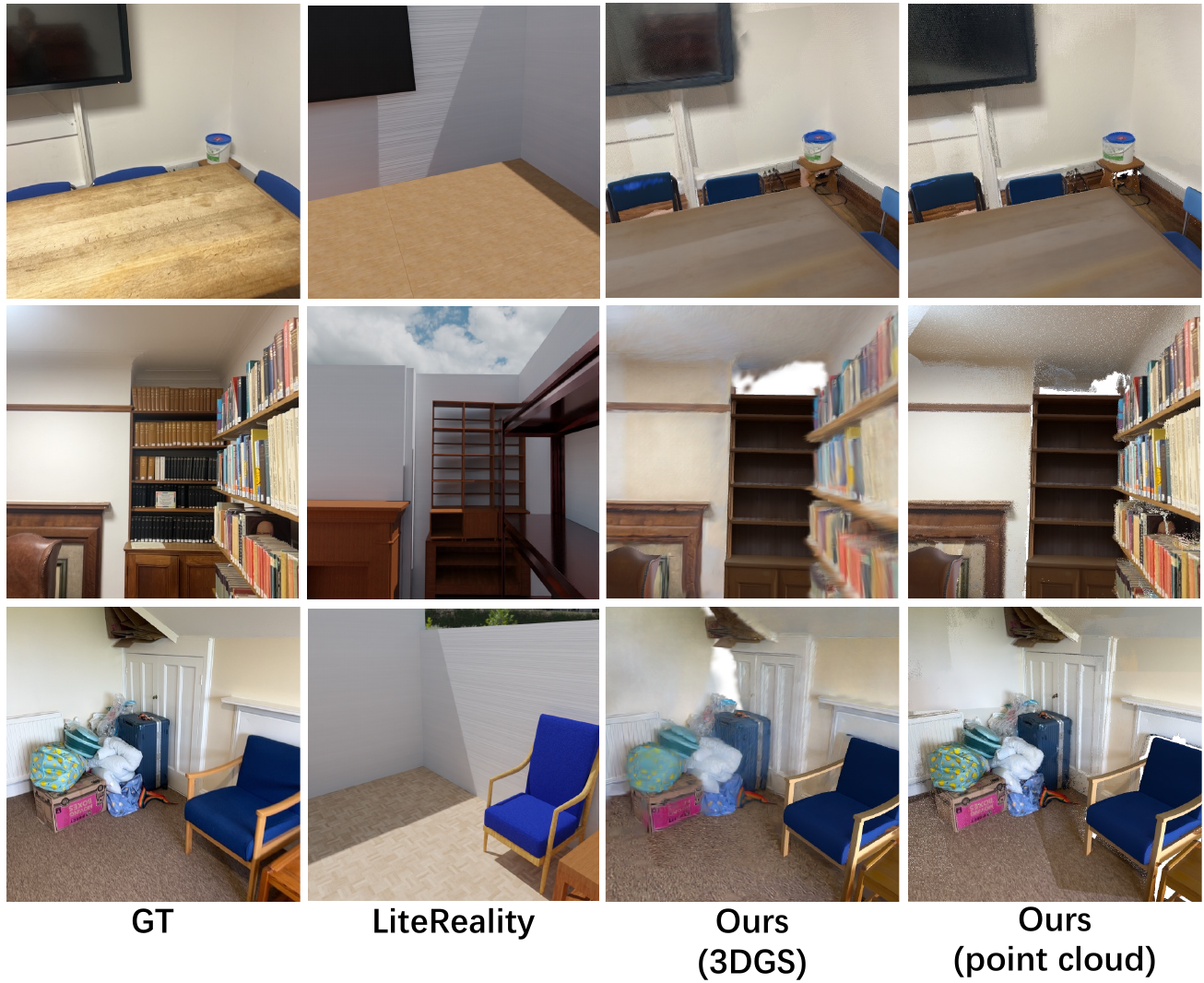}
    \caption{\textbf{Qualitative comparison with LiteReality~\cite{litereality}.} Compared with LiteReality, \methodname preserves scene-faithful background geometry and better maintains original object structure and placement.}
    \label{fig:sim_compare}
\end{figure}

\begin{figure}[t]
    \centering
    \includegraphics[width=0.9\linewidth]{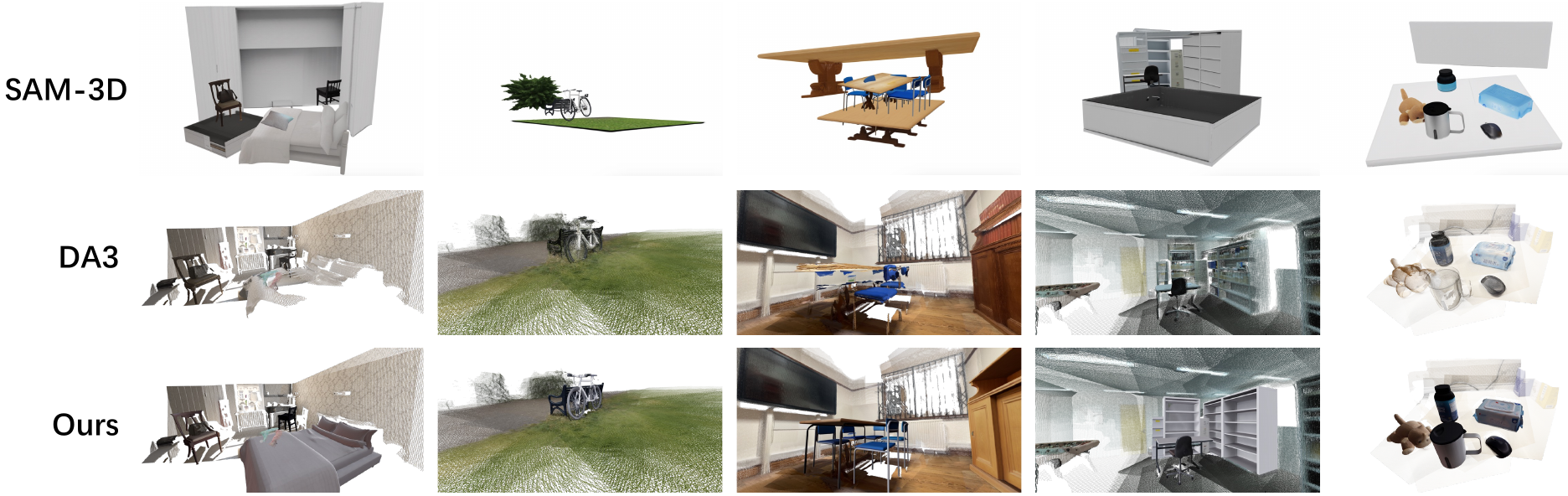}
    \caption{\textbf{Cross-dataset qualitative comparison against SAM3D~\cite{sam3d} and DepthAnything3~\cite{da3}.} Across Hypersim~\cite{hypersim}, Mip-NeRF360~\cite{mipnerf360}, LiteReality~\cite{litereality}, ETH3D~\cite{eth3d}, and our own tabletop captures, \methodname produces more complete geometry, higher-quality object reconstruction, and more consistent object layout while preserving high-fidelity background context. Our results are from post-simulation states.}
    \label{fig:qualitive}
\end{figure}

\cref{fig:sim_compare} compares \methodname with LiteReality~\cite{litereality} using the input view, LiteReality, and our two rendering variants (Ours (3DGS background) and Ours (point cloud background)). LiteReality tends to replace scene content with retrieved assets and does not preserve the original background context, which can also alter the native object structure. In contrast, our physically-grounded factorization keeps a pristine scene-faithful background while preserving scene-specific objects and their layout, yielding more coherent and interaction-ready reconstructions in both background representations.

\cref{fig:qualitive} further compares our method with DepthAnything3~\cite{da3} and SAM-3D~\cite{sam3d} across diverse domains, including Hypersim~\cite{hypersim}, Mip-NeRF360~\cite{mipnerf360}, LiteReality~\cite{litereality}, ETH3D~\cite{eth3d}, and our own tabletop-object captures. 
DepthAnything3 fails to fully reconstruct key objects, whereas SAM-3D—constrained by single-view perspectives—struggles with occluded structures, generates incorrect object geometries, and produces unreliable layout estimates with poor background fidelity.
Our method consistently recovers more complete geometry and higher-quality objects, while simultaneously preserving background fidelity and restoring physically plausible object layouts. 
% These gains are consistent with the design of \methodname: gravity-aligned global coordinates improve placement consistency, explicit mesh extraction preserves object identity, and conditional background disentanglement removes duplicated geometry without sacrificing surrounding scene details.

\subsubsection{Computational Efficiency}

\cref{tab:efficiency} shows that \methodname achieves substantially lower inference latency than LiteReality~\cite{litereality} across all five scenes. The average runtime drops from 4332.8s to 560.2s (approximately $7.7\times$ speedup), demonstrating practical efficiency gains for simulation-ready scene generation.

\begin{table}[t]
    \centering
    \caption{\textbf{End-to-end inference latency.} Runtime comparison between LiteReality~\cite{litereality} and \methodname on a unified hardware platform. We exclude evaluation-time rendering overhead.}
    \label{tab:efficiency}
    \begin{tabular}{l|cc}
        \toprule
        Scenes & LiteReality~\cite{litereality} & \textbf{Ours} \\
        \midrule
        BoardRoom-CUED             & 7686s & \textbf{1244s} \\
        Darwin-BedRoom             & 2781s & \textbf{321s} \\
        Girton-Common-Room         & 4507s & \textbf{512s} \\
        Girton-large-study-room    & 2308s & \textbf{318s} \\
        Girton-Study-Room          & 4382s & \textbf{406s} \\
        Mean                       & 4332.8s & \textbf{560.2s} \\
        \bottomrule
    \end{tabular}
\end{table}

\section{Conclusion}
\label{sec:discuss}
This paper introduced \methodname, an RGB reconstruction framework that reconciles structural interactability with geometric fidelity via physically-grounded factorization. By leveraging gravity to resolve orientation ambiguity, our pipeline enables efficient user-prompted 6-DoF object anchoring and background disentanglement within a structured hybrid representation.
\methodname enhances both object and scene-level fidelity while achieving a $7.7\times$ latency reduction compared to retrieval-based baselines. This formulation offers a practical, scalable framework for generating interactive digital twins in embodied AI.

% \clearpage\mbox{}Page \thepage\ of the manuscript.
% \clearpage\mbox{}Page \thepage\ of the manuscript.
% \clearpage\mbox{}Page \thepage\ of the manuscript.
% \clearpage\mbox{}Page \thepage\ of the manuscript.
% \clearpage\mbox{}Page \thepage\ of the manuscript. This is the last page.
% Now we have reached the maximum length of an ECCV \ECCVyear{} submission (excluding references and acknowledgements).
% References should start immediately after the main text, but can continue past p.\ 14 if needed. 
% \clearpage  % TODO FINAL: This \clearpage needs to be removed from both review and camera-ready versions.

\section*{Acknowledgements}
This work is supported by the National Key R\&D Program of China (No. 2024YFB3909903).

\bibliographystyle{splncs04}
\bibliography{main}

\clearpage
\appendix
% Keep supplementary hyperref anchors distinct from the main-paper sections.
\renewcommand{\theHsection}{supplement.\arabic{section}}
\renewcommand{\theHsubsection}{supplement.\arabic{section}.\arabic{subsection}}
\begin{center}
    \textbf{\Large Supplementary Material}
\end{center}
\vspace{-0.75em}
\section*{Overview}

This supplementary material provides additional technical details and experimental results for \methodname. \Cref{sec:supp-gv-architecture} describes the architecture of the Gravity-View (GV) alignment module. \Cref{sec:supp-point-disentanglement} elaborates on the point-conditional background disentanglement pipeline, including the construction of training data and the implementation of simulation-to-reality augmentation. \Cref{sec:supp-pointcloud-eval} explains the point-cloud evaluation strategy utilized in our NVS ablation studies. \Cref{sec:supp-vlm-target-selection} describes VLM automatic target selection as a plug-in interface that directly replaces manual target selection. \Cref{sec:supp-flow-video} presents a flowchart of the interactive workflow and introduces the accompanying supplementary video, which demonstrates the complete pipeline, additional qualitative results with multi-view renderings, and physics simulation. \Cref{sec:supp-qualitative} provides extended qualitative comparisons against SAM-3D~\cite{sam3d} and DepthAnything3~\cite{da3}. \Cref{sec:supp-failure} discusses representative failure cases.

\FloatBarrier

\section{GV Network Architecture}
\label{sec:supp-gv-architecture}

\begin{figure}[t]
    \centering
    \includegraphics[width=0.95\linewidth]{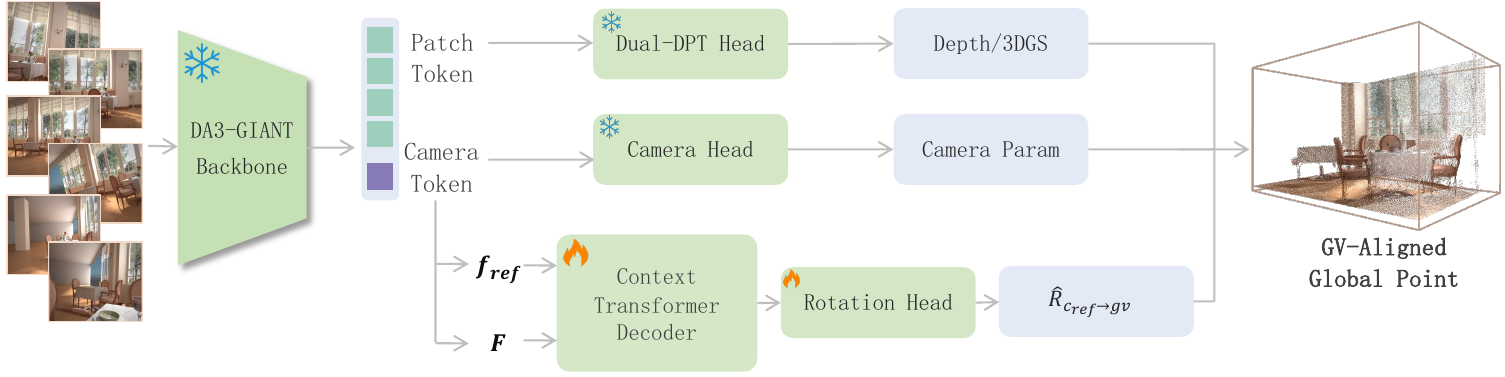}
    \caption{\textbf{Architecture of the GV prediction branch.} Multi-view images are processed by the shared DA3-Giant backbone to obtain view-wise global tokens. The token of the selected reference view is used as the query, and the tokens from all views are aggregated by a lightweight transformer decoder. The decoded feature is finally regressed to a 6D rotation representation and converted into the GV rotation matrix.}
    \label{fig:gv-network}
\end{figure}

Our GV prediction branch is built on top of the DepthAnything3-Giant-1.1~\cite{da3} model. As shown in \cref{fig:gv-network}, this section provides architecture details that complement the main-paper formulation.

\subsection{Backbone Features and Reference-View Token}
\label{sec:supp-gv-backbone}

We use the Giant version of DinoV2 as the shared multi-view backbone. Following the DA3 setup, the backbone uses a patch size of $14$ and exposes multi-scale features from layers $\{19, 27, 33, 39\}$. The final global token associated with each input view has dimension $3072$. Input RGB images are first normalized with ImageNet statistics and then processed jointly across all views. During GV training, the backbone is frozen, which preserves the strong geometric prior already learned by the large-scale reconstruction model while keeping the additional GV branch lightweight.

For a mini-batch with $B$ scenes and $N$ input views, the backbone produces a camera-token tensor $\mathbf{F} \in \mathbb{R}^{B \times N \times 3072}$. The backbone also returns a dynamically selected reference-view index for each scene using the same reference-view selection strategy as DA3. Denoting this index by $r$, we gather the corresponding reference token $\mathbf{f}_{ref} = \mathbf{F}[:, r, :] \in \mathbb{R}^{B \times 3072}$, which serves as the query token for GV decoding.

\subsection{Context Transformer Decoder}
\label{sec:supp-gv-decoder}

To aggregate gravity-related cues from all input images, we attach a lightweight transformer decoder to the final-layer camera tokens. The selected reference token $\mathbf{f}_{ref}$ is used as the query, while the full token sequence $\mathbf{F}$ from all views serves as contextual memory. Both branches are first projected from dimension $3072$ to a decoder space of dimension $512$. The decoder then updates the reference token through $5$ transformer blocks with $8$ attention heads and outputs a $1536$-dimensional feature.

Each block consists of self-attention on the reference token, cross-attention from the reference token to the multi-view context, and a feed-forward MLP, all with standard residual connections. Rotary positional encoding is used in attention. The full decoder can be summarized as
\[
\mathbf{h}_0 = \phi_x(\mathbf{f}_{ref}), \quad
\mathbf{C} = \phi_y(\mathbf{F}), \quad
\mathbf{h}_{l+1} = \mathcal{B}(\mathbf{h}_l, \mathbf{C}), \quad
\mathbf{z}_{gv} = \psi(\mathbf{h}_5),
\]
where $\phi_x(\cdot)$ and $\phi_y(\cdot)$ denote the input projections for the query and context branches, respectively, and $\psi(\cdot)$ denotes the output projection.

This decoder aggregates multi-view structural evidence such as vertical walls, floor-wall intersections, and other cross-view geometric regularities.

\subsection{Rotation Head and Coordinate Conversion}
\label{sec:supp-gv-head}

The decoded feature is processed by a lightweight MLP head consisting of two linear layers with a GELU activation function in between, following the architecture $1536 \rightarrow 512 \rightarrow 6$. The final $6$-dimensional vector is interpreted as a continuous 6D rotation representation and is subsequently orthonormalized to produce a valid $3 \times 3$ rotation matrix, yielding $\hat{\mathbf{R}}_{c_{ref}\rightarrow gv} \in \mathbb{R}^{3 \times 3}$.

Within the complete DA3-based pipeline, camera extrinsics are concurrently estimated by the original DA3 camera decoder. When these extrinsics are available, the GV branch integrates the reference-view prediction with the estimated camera rotation to derive the transformation from the canonical camera-0 frame to the GV space, denoted as $\hat{\mathbf{R}}_{c_0\rightarrow gv} = \hat{\mathbf{R}}_{c_{ref}\rightarrow gv}\hat{\mathbf{R}}_{c_0\rightarrow ref}$.
In this formulation, $\hat{\mathbf{R}}_{c_0\rightarrow ref}$ is extracted from the predicted camera rotation corresponding to the selected reference view.

\subsection{Training Objective and Implementation Details}
\label{sec:supp-gv-training}

The supervision target is the ground-truth GV rotation of the selected reference view, stored in the batch as $\mathbf{R}^{gt}_{c\rightarrow gv}$. Because the reference view is dynamically chosen, we gather the corresponding ground-truth matrix with the predicted reference index and apply the rotation loss defined in the main paper.

As described in the main manuscript, the GV alignment module is trained for $10$k steps on TartanAir~\cite{tartanair}, Hypersim~\cite{hypersim}, and vKitti~\cite{vkitti} datasets using the AdamW optimizer, with a peak learning rate of $5\times10^{-6}$ and a $1$k-step linear warm-up schedule. In practice, the GV-specific learnable parameters are highly compact: beyond the frozen multi-view backbone, the architecture only incorporates the transformer decoder and the final rotation regression head. Consequently, the GV branch is easily integrated into the complete DA3 reconstruction framework without substantially increasing the computational cost during training or inference.

\section{Point-Conditional Background Disentanglement Details}
\label{sec:supp-point-disentanglement}

\subsection{Training Data Construction from InternScenes-Real2Sim}
\label{sec:supp-point-disentanglement-data}

We construct the training data utilizing the Real2Sim subset of InternScenes~\cite{internscenes}, which comprises four source datasets: ScanNet~\cite{scannet}, Matterport3D~\cite{matterport3d}, ARKitScenes~\cite{arkitscenes}, and 3RScan~\cite{3rscan}. For each scene, the provided layout annotations and retrieved asset bindings are employed to reconstruct a structured 3D environment. Specifically, each object mesh is loaded from the asset library and transformed into a canonical object coordinate frame. The mesh is subsequently scaled to align with the dimensions of the annotated bounding box, rotated according to the layout parameters, and translated to the target centroid. Finally, the mesh is transformed into a unified scene coordinate system, where the vertical axis is oriented along the $+y$ direction to ensure consistency with our gravity-view (GV) frame. Furthermore, structural meshes for the floor, walls, and ceiling are incorporated when available, ensuring that the resulting composed scene provides both detailed object geometry and relevant proximal supporting context.

Leveraging the reconstructed scenes, we generate object-scene training pairs for conditional point classification. For each annotated instance, we first filter out invalid or unsuitable cases. These include instances characterized by missing retrieved assets, extreme dimensions, or degenerate bounding boxes, as well as specific categories unsuitable for the disentanglement of rigid objects, such as windows, curtains, mirrors, and clothing. For every remaining instance, we extract the standalone mesh to serve as the target object and generate a local scene crop by randomly expanding the corresponding bounding box. Subsequently, we aggregate all scene triangles that intersect this expanded 3D region and sample points from these surfaces with probabilities proportional to the area of the respective triangles. A sampled point is assigned a positive label if it originates from the target instance and a negative label otherwise. Consequently, each training sample consists of an object mesh accompanied by a local point cloud that provides precise binary supervision for the object and the background.

\subsection{Artifact-Oriented Training Augmentations}
\label{sec:supp-point-disentanglement-aug}

Although the supervision generated from InternScenes-Real2Sim~\cite{internscenes} is geometrically exact, a clear domain gap remains: simulated point clouds are clean, while real multi-view reconstructions contain systematic artifacts. To improve transfer to real scenes while preserving nearby background geometry during disentanglement, we apply four artifact-oriented 3D augmentations during training:
\begin{itemize}
    \item \textbf{Multi-layer Ghosting:} introduces spatially coherent duplicate layers and drift offsets to mimic multi-view fusion ghosts.
    \item \textbf{Thin-part Deformation:} perturbs slender structures (e.g., chair legs and poles) to simulate topology distortion in reconstructed geometry.
    \item \textbf{Virtual Multi-View Scanning:} re-samples points from synthetic camera trajectories with z-buffer projection to reproduce realistic visibility, density, and occlusion patterns.
    \item \textbf{Mesh Pose \& Scale Jittering:} applies random global or anisotropic scaling, translation, and rotation to condition points to model uncertainty from amodal mesh generation and pose alignment.
\end{itemize}
These augmentations enhance simulation-to-reality transfer and facilitate more reliable point-level disentanglement within real reconstructed scenes. Consequently, this approach yields cleaner object removal while effectively preserving the integrity of the adjacent background geometry.

To construct training pairs, we randomly expand retrieved object bounding boxes when cropping global scene point clouds. For the artifact-oriented augmentations above, we set the base voxel size to $0.15$ m and use: (1) Multi-layer Ghosting with $5$--$12$ layers and $0.2$--$0.5$ voxel drift; (2) Thin-part Deformation with $50\%$ probability; (3) Virtual Multi-View Scanning with $25$--$45$ camera viewpoints per scene; and (4) Mesh Pose \& Scale Jittering with global scaling $[0.65, 1.35]$, anisotropic scaling $[0.85, 1.15]$, gravity-axis rotation within $\pm 20^\circ$, and translation within $\pm 0.36$ m.

\subsection{Task-Set Evaluation}
\label{sec:supp-point-disentanglement-results}

% We construct a task-specific evaluation split from the InternScenes-ScanNet subset by holding out $20\%$ of the scenes for testing. We report two settings: \textit{w/o augmentation}, where both training and evaluation use the original held-out samples, and \textit{w/ augmentation}, where the same augmentation pipeline is applied to both training and test samples.

% \begin{table}[t]
%     \centering
%     \caption{\textbf{Task-set evaluation on the held-out InternScenes-ScanNet split.} We compare the point-conditional background disentanglement module trained with and without artifact-oriented augmentations.}
%     \label{tab:supp-point-disentanglement-taskset}
%     \begin{tabular}{l|ccccc}
%         \toprule
%         Setting & F1 $\uparrow$ & IoU $\uparrow$ & Acc. $\uparrow$ & Precision $\uparrow$ & Recall $\uparrow$ \\
%         \midrule
%         w/o augmentation & 0.993 & 0.986 & 0.993 & 0.991 & 0.994 \\
%         w/ augmentation & 0.976 & 0.974 & 0.992 & 0.984 & 0.989 \\
%         \bottomrule
%     \end{tabular}
% \end{table}

% As shown in \cref{tab:supp-point-disentanglement-taskset}, both settings achieve strong performance. The \textit{w/o augmentation} setting reaches F1 $0.993$ and IoU $0.986$, while the \textit{w/ augmentation} setting still maintains high accuracy, precision, and recall under augmented test samples. This suggests that the model remains robust when evaluated on more challenging perturbed inputs.

We construct a task-specific evaluation split from the InternScenes-ScanNet subset by holding out 20\% of the scenes for testing. 
The proposed point-conditional background disentanglement module demonstrates high efficacy and robustness on this held-out split. In the standard evaluation setting, which excludes artifact-oriented augmentations, the module achieves near-perfect performance with an F1 score of 0.993, an IoU of 0.986, and an accuracy of 0.993, while maintaining a precision of 0.991 and a recall of 0.994. To further assess the stability of the module, we evaluate its performance under more rigorous conditions by introducing artifact-oriented augmentations to the test samples. Despite the increased complexity of these perturbed inputs, the module sustains strong performance, yielding an F1 score of 0.976, an IoU of 0.974, and an accuracy of 0.992, with precision and recall remaining at 0.984 and 0.989, respectively. These results indicate that the proposed model is not only effective on clean data but also highly resilient to common point cloud artifacts.
\section{Point-Cloud Evaluation Strategy for Ablation}
\label{sec:supp-pointcloud-eval}

In the point-cloud ablation study described in the main manuscript, we evaluate geometric quality exclusively within an object-centric local region rather than across the entire scene. This design aligns with the discussion presented in the main manuscript: under the constrained viewpoints characteristic of the novel view synthesis (NVS) benchmark, a full-scene point-cloud evaluation is heavily influenced by under-observed regions. Consequently, such an evaluation fails to faithfully represent the quality of local scene factorization in the vicinity of the target object.

Initially, the target object region for each scene is identified and manually verified to ensure that the cropped volume consistently encompasses the target object alongside its immediate supporting context. All predicted point clouds are aligned with the ground-truth coordinate system prior to the computation of metrics. Following the standard VGGT~\cite{vggt} evaluation pipeline, the process begins by matching predicted and ground-truth camera centers, followed by the estimation of a global Sim(3) transformation via Umeyama~\cite{umeyama} alignment, and finally the application of ICP refinement.

Nearest-neighbor point-cloud distances are computed in both directions to derive accuracy and completion metrics; additionally, normal consistency is calculated following the same evaluation routine. Under this protocol, superior completeness indicates that a method reconstructs a more comprehensive local object-support region, whereas accuracy and normal consistency reflect geometric precision within the shared local crop.

\section{VLM Automatic Target Selection}
\label{sec:supp-vlm-target-selection}

\begin{figure}[ht]
    \centering
    \includegraphics[width=0.65\linewidth]{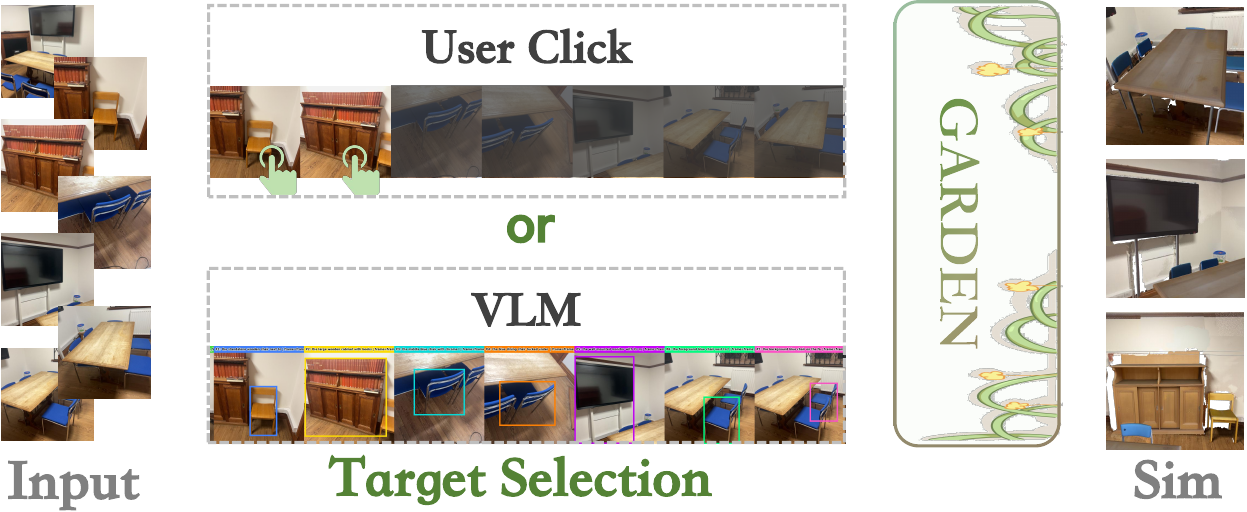}
    \caption{\textbf{VLM automatic target selection.} VLM-proposed target boxes directly replace manual user-click boxes and are passed to the same downstream GARDEN pipeline.}
    \label{fig:vlm-target-selection}
\end{figure}

The VLM automatic target-selection interface can be directly plugged into the target-selection stage, as shown in \cref{fig:vlm-target-selection}. In our implementation, Gemini-3-Flash serves as the VLM. In this mode, the manual user-click box is directly replaced by boxes generated from a two-step query procedure. First, the VLM is prompted with the multi-view observations and the downstream simulation goal, and it proposes bounding boxes for task-relevant interactive objects. Second, a critic query checks the proposed boxes against the image evidence and asks the VLM to repair obvious omissions or inaccurate boxes. The final boxes use the same interface format as manual boxes and are consumed by the unchanged downstream pipeline. This replacement automates target selection without changing object generation, pose refinement, background disentanglement, or simulation.

\section{Flowcharts and Supplementary Video}
\label{sec:supp-flow-video}

\begin{figure}[t]
    \centering
    \includegraphics[width=0.9\linewidth]{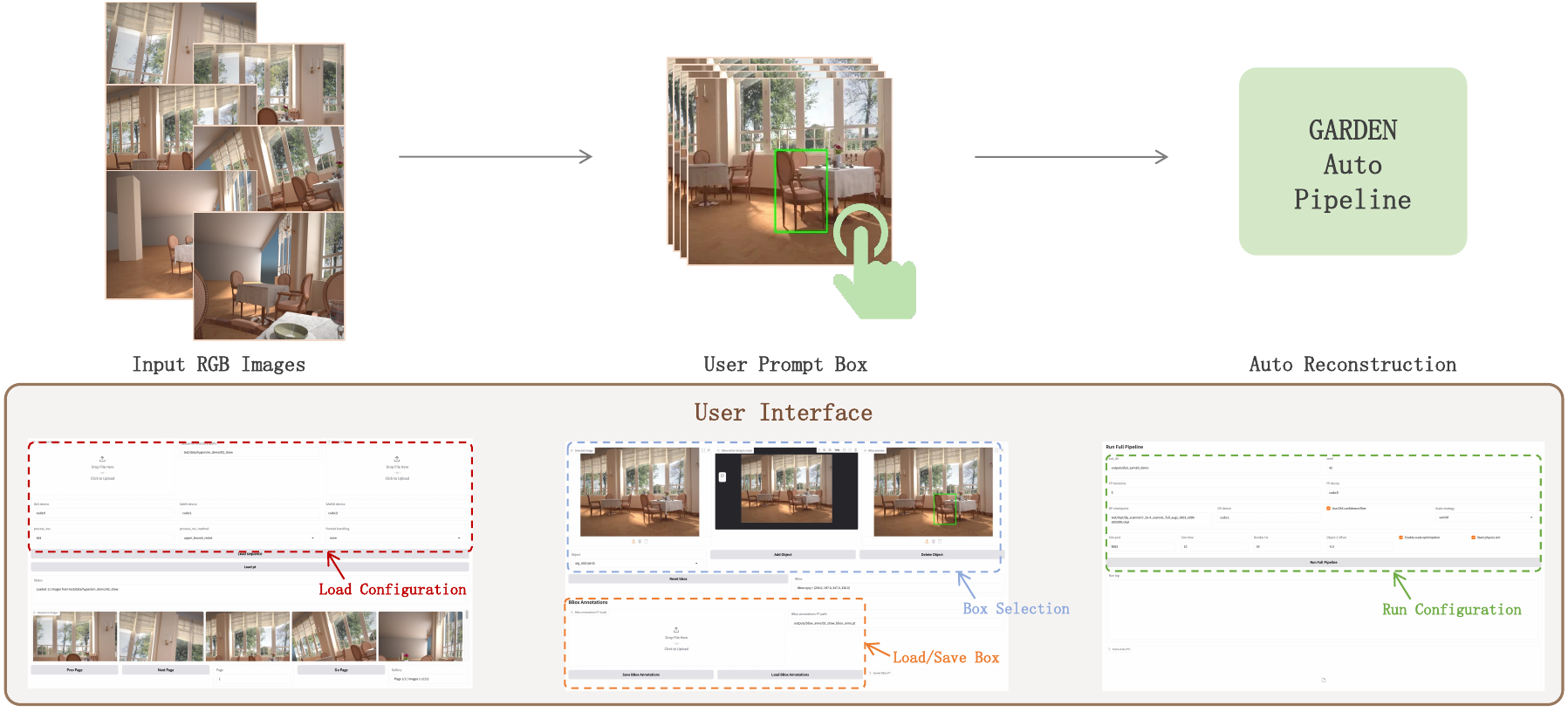}
    \caption{\textbf{Interactive workflow and user interface of \methodname.} The upper row provides a concise illustration of the target-selection pipeline, while the lower row shows the corresponding interface components in a one-to-one manner. From left to right, the workflow covers multi-view RGB input and configuration loading, optional user box selection together with image visualization and box load/save operations, and finally automatic pipeline execution with the corresponding run configuration.}
    \label{fig:flowchart}
\end{figure}

The flowchart of the interactive reconstruction procedure is provided in \cref{fig:flowchart}. Two distinct rows are presented: the upper row provides a concise illustration of the workflow, while the lower row displays the corresponding user-interface modules utilized to execute each step in practice.

In the accompanying supplementary video, we provide comprehensive visual demonstrations of our system across three dimensions. First, we present a step-by-step screen recording of the user interaction, spanning from initial configuration to automatic pipeline execution. Second, we present additional qualitative results with multi-view renderings to better illustrate the reconstructed geometry, object placement, and scene appearance from different viewpoints. Third, we demonstrate the sim-ready nature of our reconstructed scenes by showcasing physical simulations of newly introduced objects interacting with the environment.

\section{Additional Qualitative Results}
\label{sec:supp-qualitative}

We provide additional qualitative results to further illustrate the reconstruction quality and physical simulation capability of \methodname.

\begin{figure}[t]
    \centering
    \includegraphics[width=0.98\linewidth]{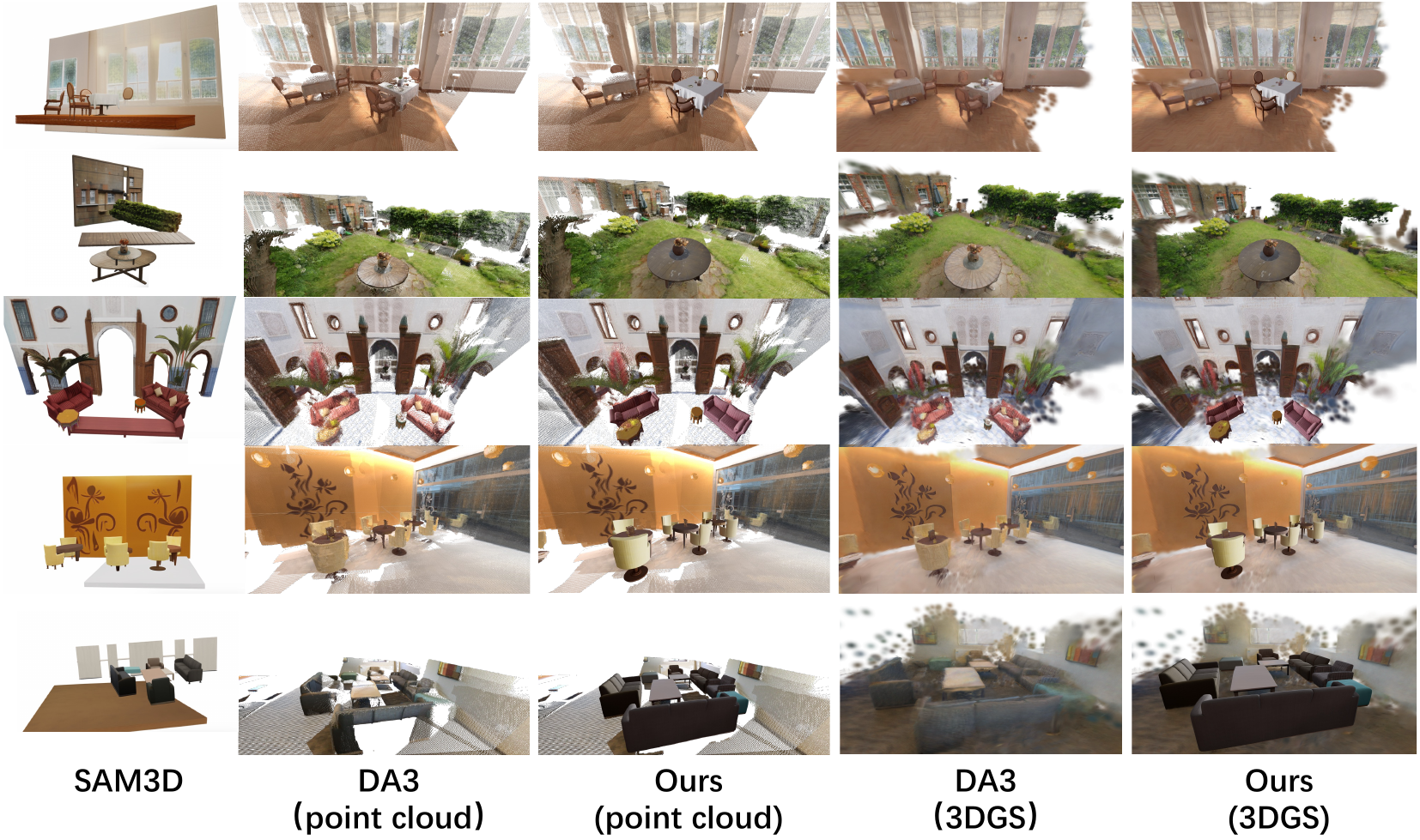}
    \caption{\textbf{Additional Qualitative comparison against SAM-3D~\cite{sam3d} and DepthAnything3~\cite{da3} on Hypersim~\cite{hypersim}, Mip-NeRF360~\cite{mipnerf360} and ScanNet~\cite{scannet} Datasets.} All our results are from post-simulation states.}
    \label{fig:sup_qualitative}
\end{figure}

\cref{fig:sup_qualitative} illustrates extended qualitative comparisons on Hypersim~\cite{hypersim}, Mip-NeRF360~\cite{mipnerf360} and ScanNet~\cite{scannet} Datasets. Each row corresponds to a single scene, while the columns display the results of SAM-3D~\cite{sam3d}, DepthAnything3~\cite{da3} (point cloud background), DepthAnything3 (3DGS background), and our method across both background representations. In comparison with these baselines, \methodname preserves scene-faithful background geometry and more effectively maintains the original object structure and spatial placement.

\begin{figure}[t]
    \centering
    \includegraphics[width=0.5\linewidth]{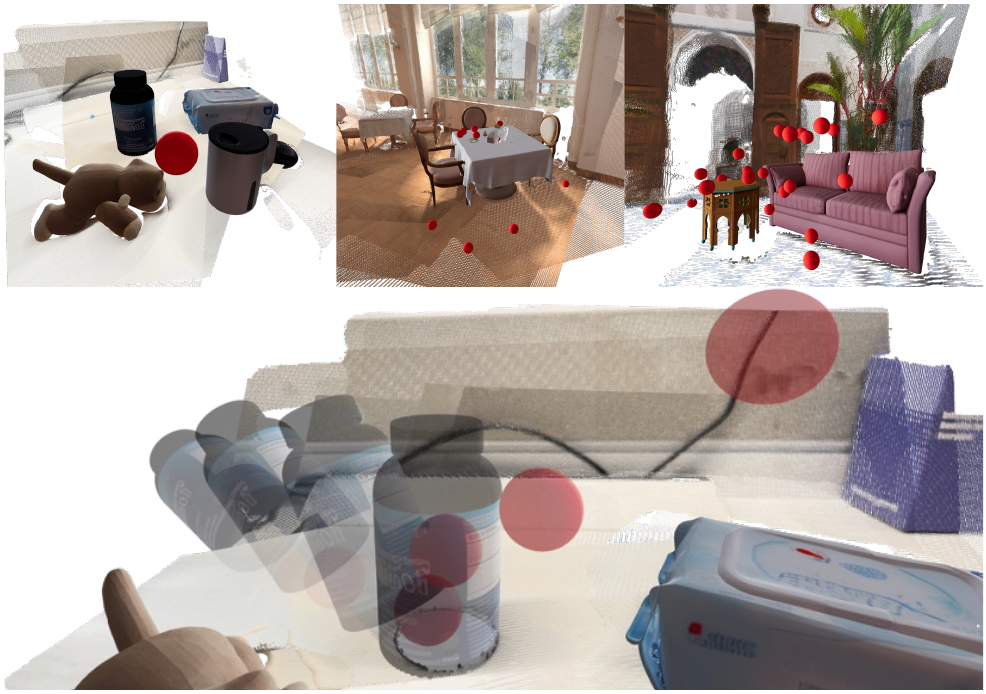}
    \caption{\textbf{Physics simulation in gravity-aligned reconstructions.} One frame from the simulation process, with rigid-body spheres (red) interacting in our reconstructed environments. The supplementary video presents the complete simulation results.}
    \label{fig:phy}
\end{figure}

\cref{fig:phy} illustrates a single frame extracted from the physics simulation sequence of the reconstructed scenes. The structured hybrid representation developed in this work facilitates direct downstream deployment. Specifically, the decoupled rigid bodies and background can be integrated into MuJoCo~\cite{mujoco} for physical interaction without requiring additional processing. The complete simulation results are provided in the accompanying supplementary video.

\section{Failure Cases}
\label{sec:supp-failure}

We summarize concrete failure cases that arise from the current limits of the upstream foundation models and from insufficient visual evidence in the input images.

\paragraph{Object generation failures.}
Severe occlusion and tiny object scales can degrade mesh quality. In cluttered composite objects, SAM-3D may reconstruct only the dominant semantic subject instead of all fine-grained parts. SAM-3 can also be affected by motion blur, which limits the quality of the target mask passed to the 3D generation stage.

\paragraph{Pose estimation failures.}
Limited pixel observations can yield inaccurate object poses. When the target occupies only a very small image region, pose refinement has insufficient visual evidence to recover a reliable 6-DoF placement.

These limitations do not change the modular nature of the pipeline: improved segmentation, 3D generation, or pose estimation modules can be incorporated as plug-and-play replacements.

\end{document}